\documentclass[10 pt, twocolumn] {article}
\pdfoutput=1

\usepackage{graphics} 
\usepackage{epsfig} 
\usepackage{mathptmx} 
\usepackage{times} 
\usepackage{amsmath} 
\usepackage{amssymb}  

\usepackage{verbatim}
\usepackage{caption} 
\title{\LARGE \bf
Unsupervised Domain Alignment to Mitigate Low Level Dataset Biases
}
\author{
  Kirthi S. Sivamani\\
  Purdue University\\
  West Lafayette, IN 47906 \\
  \texttt{ksivaman@purdue.edu} \\
}

\begin{document}
\maketitle
\thispagestyle{empty}
\pagestyle{empty}

\begin{abstract}
\textit{Dataset bias is a well-known problem in the field of computer vision. The presence of implicit bias in any image collection hinders a model trained and validated on a particular dataset to yield similar accuracies when tested on other datasets. In this paper, we propose a novel debiasing technique to reduce the effects of a biased training dataset.  Our goal is to augment the training data using a generative network by learning a non-linear mapping from the source domain (training set) to the target domain (testing set) while retaining training set labels. The cycle consistency loss and adversarial loss for generative adversarial networks are used to learn the mapping. A structured similarity index (SSIM) loss is used to enforce label retention while augmenting the training set. Our methods and hypotheses are supported by quantitative comparisons with prior debiasing techniques. These comparisons showcase the superiority of our method and its potential to mitigate the effects of dataset bias during the inference stage. }
\end{abstract}
\section{Introduction}

\quad Machine learning is inductive in nature. Computer vision models try to generalize to the entire visual world by looking at just a sample of it, by means of a labeled dataset. However, the process is not of much use if the model cannot be used on unseen datasets and distributions to yield results of the same quality. Large and labeled datasets are an integral part of successful supervised machine learning algorithms. Bigger datasets lead to more general models and tend to overfit the least. More variability in the dataset allows a machine learning model to generalize better during inference. However, each dataset has an implicit bias associated with it. Models trained on these datasets reflect these biases during inference. It has been shown by Torralba et al. [19] that it is possible to train a model to learn the biases of a dataset and identify it based on a sample test image. A vanilla support vector machine (SVM) classifier can be trained for this task to achieve accuracies that far exceed the chances of a trivial guess. The dark diagonal of the confusion matrix for the classifier [36] provides substantial evidence for this. The cross generalization test [19] shows models' accuracy drop when tested on data from different distributions. These results imply that the dataset being used to train a model has an important effect on the model's representation of the visual world. This problem is known as dataset bias and the process of mitigating it is known as domain adaptation (DA). 

\quad Broadly classifying, there can be two types of dataset biases for computer vision: Low-level and High-level. Low-level dataset biases are due to the difference in color, saturation, hue, noise, resolution, and neighborhood pixel arrangement. These are the major source of bottleneck in cross domain inference accuracy when the source domain is similar in distribution to the target domain. High-level biases are those due to major differences in distribution. These include object geometry variations, types of output classes etc. In this paper, we address the former low-level biases. This is because low-level biases are persistent in any kind of domain adaptation task and high-level biases can mostly be eradicated by choosing the correct dataset as a source domain.  

\quad In this paper, we propose a novel technique to improve the current results of the cross dataset generalization test [19] for image classification and object detection. We formulate the task of domain adaptation as an unsupervised one-to-one image translation problem [9]. We employ a generative adversarial network (GAN) [1] to translate all images in the source domain to an intermediate domain. This intermediate domain is a representation of the source domain in the space of the target domain and is much closer in distribution to the target domain. The image visualizations in Figure 1 show the similarity of images in the target and intermediate domains, compared to the vastly different source domain. 

\quad The generative network learns a mapping from the source domain X to the target domain Y and uses this learnt mapping to transform the images in the source domain. This network is essentially a data augmentation network where the entire training set is augmented before training. Using this approach, we can easily decouple the image classification/object detection task from the domain adaptation task. This makes our method easily fusible with an existing computer vision pipeline to achieve better cross-domain performance.

\quad The ultimate goal of this augmentation is to replace the implicit biases of source domain with that of the target domain and have this domain shift reflect in the cross generalization test accuracy. A notable fact is that using this approach, labeled examples from the target domain are not required. As a result, it is key to ensure label retention during the translation. In essence, a generated image should have the same bounding box coordinates and class labels as that of the image using which it was generated. It is also critical for the learnt mapping to be strictly one-to one and invertible so that generated images obtained are unique and an accurate target domain representation of their corresponding inputs. These traits are achieved using 2 strong reconstruction losses: Cycle Consistency loss [9] and a novel SSIM loss for GANs. The exact formulation of these losses has been detailed in Section 3.

The main contributions of this paper have been summarized below: 
\begin{itemize}
    \item We decouple the task of domain adaptation from image classification/object detection. This makes it easy to integrate our method with any existing computer vision pipeline.
	\item We introduce a novel approach of tackling the domain adaptation problem using the concept of unsupervised image translation.
	\item We formulate the SSIM reconstruction loss in the context of Generative Adversarial Networks.
	\item We augment the training data and create a new domain of images much closer in distribution with the target domain.
	\item We conduct thorough comparisons with state of the art DA methods and qualitatively provide proof of the hypothesis behind our augmentation network.
\end{itemize}

\quad The rest of the paper is organized as follows: We first review the related works and present the background in Section 2. We then describe our approach in detail in Section 3. Section 4 conducts experiments and presents results to demonstrate the effectiveness of our method. The specifics of network architectures and hyperparameter values are also specified for reproducibility. Section 5 presents the limitations of our method and conclusions are included in Section 6.  

\begin{figure}[t]
    \includegraphics[width=1.0\linewidth]{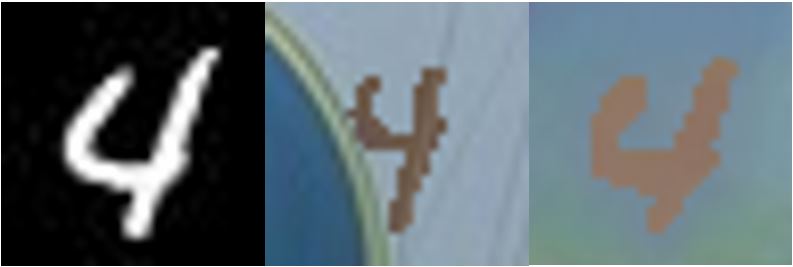}
    \caption[font=small]{Visualizations of the images from the source, intermediate, and target domains in the translation from MNIST$\rightarrow$SVHN. The leftmost image is of the digit 4 from MNIST. The middle image is of a sample image 4 from the SVHN dataset. The rightmost image is the generated image in the intermediate domain.}
    \label{fig:myplot3}
\end{figure}

\section{Related Work}

This work builds off of several popular machine learning techniques and algorithms such as generative adversarial networks, domain shift for debiasing datasets, and unsupervised domain adaptation. This section provides a review of the existing work in these fields.

\quad \textbf{Generative adversarial networks} (GANs) [1, 2] have encountered tremendous growth since their introduction in 2014. GANs are set up in the form of a minimax game between two network architectures known as the discriminator and the generator. The generator tries to generate new `fake' input samples indistinguishable from the actual input distribution. The discriminator is essentially a binary classifier that tries to determine whether a given input is real or created by the generator.  The adversarial loss [1] is used to optimize both network architectures. GANs have been used successfully for many tasks such as image generation [3], storytelling [6], representation learning [4, 5], neural style transfer [7, 8], and paired/unpaired image to image translation [9, 10].

 \quad It's the use of unpaired image to image translation
using cycle consistent adversarial networks (CycleGAN) [9] that we leverage to augment training data. The adversarial loss of CycleGAN is used for translating images from the source to intermediate domain, whereas its cycle consistency loss removes stochasticity and enforces the learnt mapping to be invertible and ensures a strict one-to-one correspondence. 

\quad This work is not the first attempt to augment data using GANs. [11] attempts to augment vanilla classifier networks by using conditional GANs to learn a data item and use it for intra-class generalization. [13] trains a CNN classifier and a CycleGAN generator to augment data for emotion classification. It employs the basic least squares loss as the adversary to resolve the vanishing gradient problem [14]. Several recent attempts at data augmentation and domain adaptation using GANs have seen encouraging results in medical applications [17, 18]. The above works do not deal with the problem of dataset bias or domain adaptation but aim to encode variance in the training set to reduce overfitting and attain better accuracy. The method we propose is the first approach to eradicate dataset bias using GAN based augmentation.

\quad \textbf{Domain shift and transfer learning} [22, 23] have been used extensively to suppress dataset bias. Many researchers have proposed methods of finding low-dimensional domain invariant features to rectify distributional mismatch between the two domains: training and testing [24, 25, 26]. Geometric deep learning methods have also yielded successful results in domain shift problems. Gong et al. [27] represent both domains in a low dimensional subspace and identify different domains as points on a Grassmann manifold. They then model the shift in the two domains using the shortest possible (geodesic) path. In later work [28], Gong et al. extend this method to use multiple geodesic flow kernels to enhance accuracy.

\quad Our work is inspired by the promising results and developments in the fields of \textbf{unsupervised domain adaptation} to mitigate dataset bias. Unsupervised methods are key as labeling data is expensive and time consuming. Ganin et al. [33] proposed the Domain Adversarial Neural Network (DANN) that learns to predict the domain of the input along with the class label using a discriminator. A gradient reversal layer is used at this discriminator to minimize source target domain discrepancy and extract domain invariant features. Adversarial Active Domain Adaptation (AADA) [37] also learns to distinguish the input image domain and uses active learning to query the most informative labels and train the DA model in a semi-supervised manner. Perhaps, the closest work to our proposal is the deep adversarial attention alignment (DAAA) [38]. DAAA uses cycle consistency loss to generate synthetic images of the source and target domain and align the obtained attention maps [35] of all resulting pairs. 

\quad Our work differs from existing work as we decouple the DA task from the classification/detection task. This approach allows us to easily integrate our augmentation network with an existing image classification or object detection pipeline. Further, our novel use of the SSIM loss gives us capability to directly transfer labels from the source domain to the intermediate domain.

\begin{figure}[t]
    \includegraphics[width=1.0\linewidth]{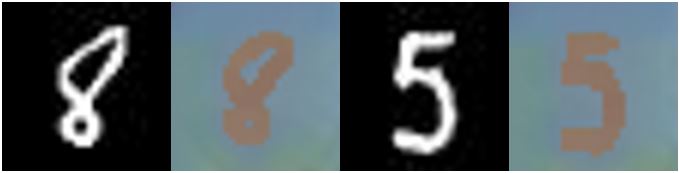}
    \caption[font=small]{The black and white images are original MNIST images of the digits 8 and 5. The corresponding images to their right are generated images in the intermediate domain for the translation from MNIST$\rightarrow$SVHN}
    \label{fig:myplot4}
\end{figure}

\section{Proposed Method}

Our approach involves decoupling the domain adaptation task from the supervised learning problem. Thus, our method can be easily blended into any independent image classification or object detection pipeline. Due to this, our method can be viewed as a means to augment source domain data as a preprocessing step for training to be better suited to the target domain. Like any trivial data augmentation method, it is key that labels from the source domain are retained after the transformation has occurred. 

\quad For this, we enforce two reconstruction losses: cycle consistency loss [9], and SSIM loss. These losses preserve underlying semantics of objects in images during translation so that class labels and bounding box coordinates can be retained. The formulation of the DA task as a one-to-one image translation problem allows the migration of low-level biases from the source to the target domain. We assume that the source and target domain distributions are not extremely distinct and that the high-level biases are not persistent. Two samples of augmented images along with their corresponding source domain inputs have been shown in Figure 2.

\quad The aim of our method is to learn a mapping between the two domains X and Y given N training samples $ [x]_i^N $ and their corresponding labels $ [y]_i^N $. The probability distribution of x is denoted as p(x) and that of y is denoted as p(y). Similar to [9], we aim to learn two mappings via two generators $G_{1}$ and $G_{2}$. $G_{1}$ takes in as input x $ \in $ X and generates fake(x) $ \in $  Y. $G_{2}$ takes in as input y $ \in $ Y and generates fake(y) $ \in $ X. Two adversarial discriminators $D_{1}$ and $D_{2}$ are trained simultaneously to distinguish the fake data generated by the generator. Our final objective function contains three losses: Adversarial loss for the training of the discriminators and generators, cycle consistency loss for the invertibility of the learnt functions and distributions, and an SSIM loss for enforcing label retention in augmented images. The generated images and labels are used to train the classifier. The losses and the network details have been elaborated in the following subsections.

\subsection{Adversarial Loss}

The adversarial loss [1] has been applied to learn both the mapping functions F: $ X \rightarrow Y $ and G: $ Y \rightarrow X $. 
The adversarial loss for the discriminator F can be expressed as: 
\begin{equation} \label{eq1}
\begin{split}
    L_{ADV} = E_{x \in p(x)}[1 - \log D_{1}(G_{1}(x))] \\
     + E_{y \in p(y)}[\log (D_{1}(y))]
\end{split}
\end{equation}

This loss $ L_{ADV} $ is structured in the form of a minimax game where $G_{1}$ tries to minimize the loss and $D_{1}$ tries to maximize the loss. $G_{1}$ generates fake images $G_{1}(x) \in Y$, and $D_{1}$ tries to distinguish between the real images y and these generated images $G_{1}(x)$. A perfect discriminator would drive the loss to a maximum of 0 by being able to distinguish the real images from the fake ones with a 100\% accuracy. The generator $G_{2}$ and discriminator $D_{2}$ are also trained using a similar loss as (\ref{eq1}) for the function of G: $Y \rightarrow X$.  

\subsection{Cycle Consistency Loss}

As pointed out by Park et al. [9], a large enough network can learn stochastic functions F and G such that it can map the same set of input images to any random permutation of images in the target domain. Any of these learnt mappings can induce an output distribution that matches the target distribution. This means that the function is not a one-to-one function and hence is not invertible. Essentially, use of adversarial losses solely cannot guarantee that a given input image i $\in$ X will be mapped to a desired output image j $\in$ Y. We employ the cycle consistency loss to a similar effect as [9] to avoid this problem. This loss ensures that both F and G are invertible and that an image in any domain can be mapped back to itself after successive translations via F and G, or vice versa. The loss is formulated as:

\begin{equation} \label{eq2}
\begin{split}
    L_{cyc}(G, F) = \lambda (E_{x \in p(x)} (x) [F(G(x)) - x] \\
    + E_{y \in p(y)} (y) [F(G(y)) - y]),
\end{split}
\end{equation}

where $\lambda$ is a hyperparameter that controls the influence of the cycle consistency loss.

\begin{table*} [t]
\caption{Cross testing results (Classification accuracy \%) on the Office-31 Dataset with ResNet-50 base architecture.}

\begin{tabular} [b] {|p{1.75cm}|p{1.65cm}|p{1.65cm}|p{1.65cm}|p{1.65cm}|p{1.65cm}|p{1.65cm}|p{1.55cm}|}
 \hline
 Method & A $\rightarrow$ D &A $\rightarrow$ W&D $\rightarrow$ A&D $\rightarrow$ W&W $\rightarrow$ A&W $\rightarrow$ D& Average\\
 \hline
 Source & 68.9\% & 68.4\% & 62.5\% & 96.7\% & 60.7\% & 99.3\% & 76.1\%\\
 \hline
 DAN [39] & 78.6\% & 80.5\% & 63.6\% & 97.1\% & 62.8\% & 99.6\% & 80.4\%\\
 \hline
 DANN [33] & 79.7\% & 82.0\% & 68.2\% & 96.9\% & 67.4\% & 99.1\% & 82.2\%\\
 \hline
 AADA [37] & 77.8\% & 86.2\% & 69.5\% & 96.2\% & 68.9\% & 98.4\% & 82.9\%\\
 \hline
 DAAA [38] & 88.8\% & 86.8\% & 74.3\% & \textbf{99.3\%} & 73.9\% & 99.9\% & 87.2\%\\
 \hline
 GTA [40] & 87.7\% & 89.5\% & 72.8\% & 97.9\% & 71.4\% & 99.8\% & 86.5\%\\
 \hline
 CDAN [41] & 93.4\% & \textbf{93.1\%} & 71.0\% & 98.6\% & 70.3\% & \textbf{99.9\%} & 87.7\%\\
 \hline
 Our method & \textbf{94.3\%} & 91.1\% & \textbf{76.0\%} & \textbf{99.3\%} & \textbf{74.0\%} & 98.9\% & \textbf{88.9\%} \\
 \hline
\end{tabular}
\label{t3}
\end{table*}

\subsection{SSIM Loss}

The Structural Similarity Index is a measure of similarity between two images. The bounded output values of SSIM between -1 for perfect mismatch and 1 for a perfect match between two images make it a good loss function.

The cycle consistency loss serves as a powerful reconstruction loss for image to image translation. However, for the purpose of data augmentation, the generated images should have sufficient similarity to their corresponding inputs so that the labels, i.e. the class label and the bounding box coordinates can be retained with a high accuracy. To ensure this label retention, we enforce the SSIM reconstruction loss on both sets of functions F and G. The formulation of SSIM as an image restoration loss has been shown in [31].

\quad Due to the powerful regularizing nature of the SSIM loss, we introduce a hyperparameter $\lambda_ {SSIM}$ to check the influence of this loss on the generation of images in the intermediate domain. Thus, the SSIM loss for two images is formulated as:

\begin{equation} \label{eq3}
\begin{split}
    L_{SSIM}(P) = \lambda _{SSIM} (\sum _{p \in P} (1 - SSIM(p))),
\end{split}
\end{equation}

where P is a pair of images {x, z} where x $\in$ X and z $\in$ Z, p is a pixel location $\in$ P, and SSIM(p) the SSIM value for the pixel location p in the images. Here, X is the source domain and Z is the intermediate domain\footnote{The exact formulation and mathematical details of SSIM(p) have been discussed in detail in [31]. Thus, we omit this discussion here. }.

\quad The SSIM loss was not included in the preliminary experiments. Evaluations were conducted with a higher value of $\lambda$ for greater emphasis on the reconstruction. However, as shown in section 4.4, the accuracy of intermediate domain labels without the SSIM loss was sub-par. The cycle consistency loss essentially makes the model pipeline similar to that of an autoencoder where the intermediate domain represent the condensed code. We hypothesize that a higher value of lambda makes the model more conservative. Consequently, the code representation is not distinct enough from the input distribution to carry out the domain shift effectively. Thus, the SSIM loss was added to the objective and satisfactory empirical results were obtained (section 4.4). 

\subsection{The Complete Objective}

Our final objective, $L_{TOTAL}$, for training the augmentation network is:

\begin{equation} \label{eq4}
\begin{split}
    L_{TOTAL} = \lambda _{SSIM} (\sum _{p \in P} (1 - SSIM(p))
    + \sum _{q \in Q} (1 - SSIM(q))) \\
    +  \lambda (E_{x \in p(x)} (x) [F(G(x)) - x]
    +  E_{y \in p(y)} (y) [F(G(y)) - y]) \\ 
    +  E_{x \in p(x)}[1 - \log D_{1}(G_{1}(x))]
     +  E_{y \in p(y)}[\log (D_{1}(y))]
\end{split}
\end{equation}

Here, P is the set of images $ \in (x, G_{1}(x))$ and Q is the set of images $ \in (y, G_{2}(y))$. $\lambda$ and $\lambda_{SSIM}$ control the relative weights of the 3 losses. 

\subsection{Training methodology}

During training, we assume that labeled samples from the source domain X and unlabeled samples from the target domain Y are available. Our data augmentation network is first trained by providing the images from the source domain and target domain training data. These domains must represent different datasets for the translation to be meaningful. The generated images in the intermediate domain are then used for training the classifier or object detector. The object class and bounding box labels for the generated images are the same as that of their corresponding inputs. `Domain alignment' refers to the transformation of the source domain to the intermediate domain.

\begin{table*} [t]
\caption{Cross testing results (person detection mAP (\%)). The datasets along the column are the training sets and those along the row are the testing set.}

\begin{tabular} [b] {|p{1.75cm}|p{1.60cm}|p{1.60cm}|p{1.60cm}|p{1.60cm}|p{1.60cm}|p{1.75cm}|p{1.75cm}|}
 \hline
 Dataset & SUN09 & LabelMe & VOC2007 & ImageNet & Caltech101 & Mean others \par (original) & Mean others \par (debiased)\\
 \hline
 SUN09 & 69.6\% & 64.4\% & 60.2\% & 60.2\% & 67.7\%& 48.1\% & 63.1\%\\
 \hline
 LabelMe & 64.0\% & 66.6\% & 60.6\% & 60.9\% & 63.0\% & 49.5\% & 62.1\%\\
 \hline
 VOC2007 & 56.3\% & 56.3\% & 56.3\% & 56.0\% & 60.0\% & 56.0\% & 57.2\%\\
 \hline
 ImageNet & 56.0\% & 55.9\% & 54.0\% & 59.6\% & 55.0\% & 45.3\% & 55.2\%\\
 \hline
 Caltech101 & 66.1\% & 47.4\% & 41.9\% & 54.0\% & 100.0\% & 20.5\% & 52.4\%\\
 \hline
 Mean others & 60.6\% & 56.0\% & 74.3\% & 54.2\% & 61.4\% & 43.9\%& 58.0\%\\
 \hline
\end{tabular}
\label{t4}
\end{table*}

\begin{table*} [t]
\caption{Cross testing results (car detection mAP (\%)). The datasets along the column are the training sets and those along the row are the testing set.}

\begin{tabular} [b] {|p{1.75cm}|p{1.60cm}|p{1.60cm}|p{1.60cm}|p{1.60cm}|p{1.60cm}|p{1.75cm}|p{1.75cm}|}
 \hline
 Dataset & SUN09 & LabelMe & VOC2007 & ImageNet & Caltech101 & Mean others \par (original) & Mean others \par (debiased)\\
 \hline
 SUN09 & 69.8\% & 66.6\% & 66.6\% & 66.2\% & 60.0\%& 47.5\% & 64.9\%\\
 \hline
 LabelMe & 64.0\% & 67.6\% & 62.6\% & 60.9\% & 59.7\% & 48.6\% & 61.8\%\\
 \hline
 VOC2007 & 57.9\% & 61.2\% & 62.1\% & 56.2\% & 61.0\% & 55.6\% & 59.1\%\\
 \hline
 ImageNet & 54.0\% & 55.5\% & 55.5\% & 60.7\% & 60.2\% & 45.4\% & 56.3\%\\
 \hline
 Caltech101 & 23.4\% & 36.4\% & 41.0\% & 39.0\% & 100.0\% & 20.4\% & 35.0\%\\
 \hline
 Mean others & 49.8\% & 55.0\% & 56.4\% & 55.6\% & 60.2\% & 43.5\%& 55.4\%\\
 \hline
\end{tabular}
\label{t5}
\end{table*}

\subsection{Training details}

\quad The original adversarial loss proposed by Goodfellow et al. [1] has shown to be unstable during training. Thus, the generator tries to minimize $L_{ADV}$ where the discriminator tries to minimize $\log (D(G(x)))$ instead of maximizing $L_{ADV}$. We use Adam [45] for optimization with a constant learning rate of 0.001 throughout the training. The network is trained on a single TitanX GPU for 60 epochs summing to 12 hours of training. The training process is not regularized by weight decay, dropout or norm based methods. The cycle consistency loss and SSIM loss act as regularizers and any form of overfitting that occurs corresponds to more transfer of bias, which is a desirable outcome for the purpose of training data augmentation.

\quad Similar to Park et al. [9], we update the discriminator using a buffer (size=50) of previously generated images to prevent model oscillation [12]. Hyperparameters $\lambda$  and $\lambda_{SSIM}$ for the final objective are set as 10 and 0.02 respectively.  When the network has been trained, all samples in the source domain can be passed into the trained network with a batch size of 1. The generated images in this intermediate domain are now much similar to the target domain and are used to train the classifier or object detector with bounding box coordinates and image labels retained from the training set.

\subsection{Network implementation}

All experiments are conducted using PyTorch in the python programming language. Most datasets are obtained and transformed directly using torchvision and torch dataloaders. All images are resized and transformed to RGB 3 color channel 256 $\times$ 256 images as a preprocessing step. Data augmentation techniques of random rotation (10 degrees) and horizontal flipping are applied. This acts as a preliminary step in achieving lateral and rotational invariance typically missing in most modern datasets [19].

\quad For setting up the Generative Network, we adopt the network structure proposed by Johnson et al. [8] in the field of super-resolution. The network does not use pooling methods for down-sampling the layers, instead, two stride-2 convolution layers are used to reduce the dimension from 256 $\times$ 256 to 64 $\times$ 64. The architecture consists of 5 residual blocks [44]. Fractionally strided convolutions [48] with a stride of 1/2 are used for up-sampling. The convolutions are followed by batch normalization [46] and the ReLU activation [47] function in a series of transformations, apart from the last layer that uses a scaled inverse tangent activation for the output to remain in the 0 to 255 pixel intensity range. Similar to the discriminator structure in [9], we employ 70 $\times$ 70 PatchGAN [10] that provides satisfactory results with relatively few parameters.

\section{Evaluation and Results}

\subsection{Datasets}

We use three sets of datasets to present our results. The first two sets are used for image classification and the last set is used for object detection. For all image classification tasks, we use the classification accuracy as the metric of evaluation and for all object detection tasks, we use the mean average precision (mAP).

\subsubsection{Image classification}

The Office-31 [32] dataset is a popular dataset in the field of domain adaptation. Office-31 consists of 3 domains: Amazon (A), Webcam (W), and DSLR (D) with 31 classes in each domain. There are a total of 4110 images in these 3 datasets. For a fair evaluation, we obtain cross generalization classification accuracy across all 6 cross domains: A $\rightarrow$ W, W $\rightarrow$ A, A $\rightarrow$ D, D $\rightarrow$ A, D $\rightarrow$ W, and W $\rightarrow$ D.

\quad The MNIST dataset of handwritten digits [42] is also used as the source domain in another set of evaluations. MNISTM [33] and SVHN[30] datasets are used as target domains. MNIST is a significantly easier dataset and this set of evaluations proved to be much more challenging for all DA algorithms in our evaluations. The transformation from SVHN to MNISTM is also conducted.

\subsubsection{Object Detection}

For object detection, we replicate the experiment performed by Torralba et al. [19] for cross generalization of person detection and car detection. For this cross generalization test, we use SUN09 [50], LabelMe [51], PASCAL VOC 2007 [52], Caltech101 [43], and ImageNet [49]. These datasets are popular in the field of machine learning and contain ground truth annotations for object labels and bounding boxes.

\subsection{Results}

\begin{table} [t]
\caption{Cross test classification accuracy (\%) results on 0-9 digit datasets.}
\begin{tabular} [b] { |p{1.2cm}|p{1.2cm}|p{1.2cm}|p{1.2cm}| p{1.2cm}| }
 \hline
 Method & Source Target & Mnist Mnistm &Mnist Svhn&Svhn Mnistm\\
 \hline
 \multicolumn{2}{|c|}{Source only}  & 54.2\% &60.2\%&   59.2\%\\
 \hline
 \multicolumn{2}{|c|}{SA [15]}& 61.6\% & 68.5\% &65.5\%\\
 \hline
 \multicolumn{2}{|c|}{DANN [33]} &77.0\% & 81\%& 77.0\% \\
 \hline
 \multicolumn{2}{|c|}{Our method} &\textbf{77.9\%} & \textbf{85\%}&  \textbf{79.2\%}\\
 \hline
 \multicolumn{2}{|c|}{Trained on Target}    &89.0\% & 93.9\%&  89.0\%\\
 \hline
\end{tabular}
\label{t2}
\end{table}

\begin{table} [t]
\caption{Label retention percentages across various evaluation tasks.}
\begin{tabular} [b] {|p{2.6cm}|p{2cm}|p{2.2cm}|}
 \hline
 DA Task & Training images (\#) & Label accuracy ($\approx$ \%)\\
 \hline
 A$\rightarrow$D&2817&99.9\%\\
 \hline
 D$\rightarrow$W&498&99.8\%\\
 \hline
 Mnist$\rightarrow$Mnistm&10000&100.0\%\\
 \hline
 svhn$\rightarrow$Mnistm&6000&100.0\%\\
 \hline
 SUN09$\rightarrow$LabelMe&10000&99.0\%\\
 \hline
 LabelMe$\rightarrow$VOC&6000&98.4\%\\
 \hline
 VOC$\rightarrow$Caltech101&6000&99.0\%\\
 \hline
\end{tabular}
\label{t1}
\end{table}

\begin{table} [t]
\caption{Label retention percentages for object detection tasks without the inclusion of the SSIM loss in the augmentation network training objective.}
\begin{tabular} [b] {|p{2.6cm}|p{2cm}|p{2.2cm}|}
 \hline
 DA Task & Training images (\#) & Label accuracy ($\approx$ \%)\\
\hline
 SUN09$\rightarrow$LabelMe&10000&94.0\%\\
 \hline
 LabelMe$\rightarrow$VOC&6000&90.0\%\\
 \hline
 VOC$\rightarrow$Caltech101&10000&90.0\%\\
 \hline
\end{tabular}
\label{t}
\end{table}

In extensive experiments, we compare our method to prior debiasing techniques and achieve state of the art performance. All evaluations are conducted under a common setting and base architecture. Reported results in all tables differ only due to difference in the domain adaptation technique. Evaluations are reported by averaging the results of 15 successive trials during testing for a robust comparison.

\quad We use the Resnet50 [44] network as our base model in the testing of the Office-31 dataset. The results in table 1 clearly show the superiority of our approach in 4 out of 6 classification tasks. Our average classification accuracy of 88.9\% for this task is a significant improvement over the 87.7\% previously achieved by CDAN [41]. The `source' row measures the performance when no DA technique is applied for testing on the target domain. 

\quad For the task of object detection, we emulate the experiment on person and car detection performed by Torralba et al. [19] on cross dataset generalization. Like the original paper [19], we follow the method of Dalal et al. [34] for object detection. This method is based on the HOG detector followed by a linear SVM. This approach is obsolete in computer vision and ineffective compared to modern deep learning based object detection techniques [21, 29]. However, using this method allows us to directly compare our results from those of the original cross dataset generalization test [19]. It also shows the power of decoupling the DA and detection task, and how our augmentation network can be easily fused with an existing framework for image classification or object detection to yield significantly better performance.

\quad We use the SUN09 [50], LabelMe [51], PASCAL VOC 2007 [52], Caltech101 [43], and ImageNet [49] datasets. We omit the MSRC dataset [16] from the original experiment. Empirical results showed us that MSRC is a relatively easy testing set and tends to skew the accuracy results towards a deceptively high value. The results in tables 2 and 3 show the cross test generalization results using our method. The datasets along the column are the training sets and the datasets along the row are the testing sets. The `mean others' column is the detection mAP mean for testing on all datasets but self. The `mean others (original)' column [19] has been included as a baseline measure for mAP when the classifier is trained only on the source domain without any DA technique applied. The results show that our method improves cross generalization by an average of 14.1\% for person detection and 11.9\% for car detection.

\quad For the recognition of handwritten digits, we use a baseline architecture consisting of 2 convolutional layers, each followed by ReLU activation and maxpooling with a stride of 2\footnote{This architecture is taken from the PyTorch examples repository on GitHUb that uses the MNIST dataset. The code can be found at https://github.com/pytorch/examples/blob/master/mnist/main.py}. Our approach outperforms DANN [33] on all 3 DA tasks given in table 4. We also compare with unsupervised subspace alignment for DA [15] and use a strategy similar to [33] to set up the experiment. Over the 3 conducted tasks for handwritten digit recognition, we improve by an average of 2.37\% over DANN [33].

\subsection{Label Retention and SSIM loss}

We have previously emphasized the importance of retaining training labels after data augmentation. We further explore the accuracy of label retention and the role played by the SSIM loss in the same. The accuracy of retained labels and bounding boxes is measured for every newly generated training batch by manual inspection of randomly sampled images from the intermediate domain. The bounding box labels were checked for a threshold of 90\% IOU with the ground truth\footnote{The ground truth for these randomly sampled intermediate domain images were also obtained manually.} before being assigned as incorrect. This procedure is taxing but is the only way of estimating the accuracy of labels of the images in the intermediate domain.

\quad Table 5 shows the label retention accuracy for various classification and detection DA tasks. It is clear that the labels are always correct for the intermediate domain for image classification, and have a high accuracy ($\approx 99\%$) for object detection tasks as well. Table 6 shows the label retention accuracy for the object detection tasks given in Table 5, but without the use of SSIM loss in the augmentation network training objective. An accuracy of 90\% correct training labels for two detection tasks is below par, especially when the incorrectly labeled images cannot be filtered out automatically. This would likely yield mediocre detection results due to a high number of incorrect training labels. These results demonstrate the importance of the SSIM reconstruction loss, especially for the task of object detection.

\subsection{T-SNE visualization of domains}

The evaluations until this point have been quantitative and have shown the superiority of our method in several DA tasks. A more qualitative analysis would give a better intuition of our method and an understanding of the transfer of low level biases from one domain to the other. Using T-distributed stochastic neighborhood embedding (T-SNE) [20], we explore a method to visualize the source, intermediate, and target domains for two DA tasks. To obtain the visualizations, we use T-SNE to represent 100 randomly sampled images from each domain in 2 dimensions. The 2-dimensional points are then scattered on a plot, separated by the domain. We extract a 512 length feature vector for each image using the pretrained Resnet18 model [44] before using T-SNE for compression to 2 dimensions.

\quad This visualization was performed for 2 DA tasks. Figure 1 shows the generated plot for the MNIST$\rightarrow$MNISTM transformation. Figure 2 shows the same result for the adaptation from VOC to ImageNet. Both graphs show that the cluster of the points from the intermediate domain is significantly closer to the target domain. This showcases the successful transfer of certain image characteristics across the domains and confirms our hypothesis of transferring low-level biases using one-to-one image translation. 

\begin{figure}[t]
    \includegraphics[width=1.0\linewidth]{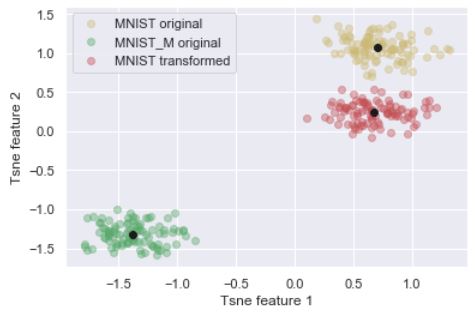}
    \caption[font=small]{T-SNE visualization of the MNIST, MNISTM, and intermediate MNIST$\rightarrow$MNISTM domains.}
    \label{fig:myplot}
\end{figure}

\begin{figure}[t]
    \includegraphics[width=1.0\linewidth]{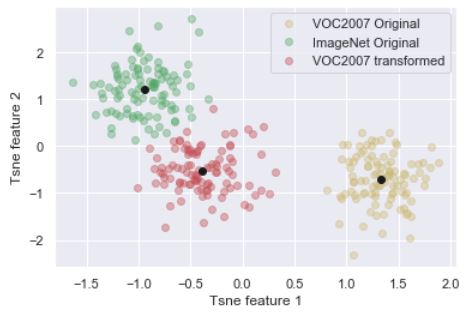}
    \caption[font=small]{T-SNE visualization of the VOC, ImageNet, and intermediate VOC$\rightarrow$ImageNet domains.}
    \label{fig:myplot2}
\end{figure}

\section{Limitations}

Our approach to DA yields state of the art results, but it does not come without shortcomings. As of today, we see two limitations to our approach: \textbf{Specificity} and \textbf{persistent high-level biases}. Specificity refers to the problem of our augmentation network being usable only for the specific domain adaptation task it has been trained for. It requires retraining every time the target domain gets changed. Consequently, it would lead to sub-optimal results in case of a shift in target domain distribution.

\quad Another drawback is that our augmentation network itself is a biased translation network regularized by heavy reconstruction losses. This makes the image to image translation conservative and does not allow cross domain transfer of high-level biases such as object geometry and class distributions. An approach towards transfer of high-level biases would be to use a different GAN for augmentation that allows shift in object location and geometry within an image. Here, one would have to be careful of the validity of training set labels for the intermediate domain. Addressing these shortcomings is key to success of future works in our approach. 

\section{Conclusion}

In this paper, we discuss how the unsupervised DA task can be formulated as a one-to-one image translation problem. We propose an augmentation network that translates the source domain to an intermediate domain, much closer in resemblance with the target domain. We use the adversarial loss, cycle consistency loss, and a novel SSIM loss for training the augmentation network. Our proposed SSIM loss was shown to be critical for accuracy of labels in the intermediate domain when retained from the source domain. Our approach is specific to the DA task, thus yielding state of the art empirical results on several tasks. Qualitative analysis using T-SNE visualization of domains demonstrated the power of our method in successfully capturing and transferring low-level dataset biases across domains. Our approach tackles the unsupervised domain adaptation task with a novel perspective, and we plan to further explore this method in the future to overcome the limitations of specificity and persistent high-level biases.

\bibliographystyle{ieeetr}
\nocite{*}
\bibliography{ref}

\begin{thebibliography}{10}

\bibitem{NIPS2014_5423}
I.~Goodfellow, J.~Pouget-Abadie, M.~Mirza, B.~Xu, D.~Warde-Farley, S.~Ozair,
  A.~Courville, and Y.~Bengio, ``Generative adversarial nets,'' in {\em
  Advances in Neural Information Processing Systems 27}, 2014.

\bibitem{Zhao2016EnergybasedGA}
J.~J. Zhao, M.~Mathieu, and Y.~LeCun, ``Energy-based generative adversarial
  network,'' in {\em ICLR}, 2017.

\bibitem{Radford2016UnsupervisedRL}
A.~Radford, L.~Metz, and S.~Chintala, ``Unsupervised representation learning
  with deep convolutional generative adversarial networks,'' in {\em ICLR},
  2016.

\bibitem{Mathieu2016DisentanglingFO}
M.~Mathieu, J.~J. Zhao, P.~Sprechmann, A.~Ramesh, and Y.~LeCun, ``Disentangling
  factors of variation in deep representations using adversarial training,'' in
  {\em Advances in Neural Information Processing Systems}, 2016.

\bibitem{Salimans2016ImprovedTF}
T.~Salimans, I.~J. Goodfellow, W.~Zaremba, V.~Cheung, A.~Radford, and X.~Chen,
  ``Improved techniques for training gans,'' in {\em Advances in Neural
  Information Processing Systems}, 2016.

\bibitem{Li2018StoryGANAS}
Y.~Li, Z.~Gan, Y.~Shen, J.~Liu, Y.~J. Cheng, Y.~Wu, L.~Carin, D.~E. Carlson,
  and J.~Gao, ``Storygan: A sequential conditional gan for story
  visualization,'' {\em ArXiv preprint}, vol.~abs/1812.02784, 2018.

\bibitem{7780634}
L.~A. {Gatys}, A.~S. {Ecker}, and M.~{Bethge}, ``Image style transfer using
  convolutional neural networks,'' in {\em CVPR}, 2016.

\bibitem{Johnson2016PerceptualLF}
J.~Johnson, A.~Alahi, and L.~Fei-Fei, ``Perceptual losses for real-time style
  transfer and super-resolution,'' {\em ECCV}, 2016.

\bibitem{CycleGAN2017}
J.-Y. Zhu, T.~Park, P.~Isola, and A.~A. Efros, ``Unpaired image-to-image
  translation using cycle-consistent adversarial networkss,'' in {\em ICCV},
  2017.

\bibitem{8100115}
P.~{Isola}, J.~{Zhu}, T.~{Zhou}, and A.~A. {Efros}, ``Image-to-image
  translation with conditional adversarial networks,'' in {\em CVPR}, 2017.

\bibitem{DAGAN}
A.~Antoniou, A.~Storkey, and H.~Edwards, ``Data augmentation generative
  adversarial networks,'' {\em ArXiv preprint}, vol.~abs/1711.04340, 2018.

\bibitem{8099724}
A.~{Shrivastava}, T.~{Pfister}, O.~{Tuzel}, J.~{Susskind}, W.~{Wang}, and
  R.~{Webb}, ``Learning from simulated and unsupervised images through
  adversarial training,'' in {\em CVPR}, 2017.

\bibitem{EC-DAGAN}
X.~Zhu, Y.~Liu, J.~Li, T.~Wan, and Z.~Qin, {\em Emotion Classification with
  Data Augmentation Using Generative Adversarial Networks}.
\newblock 2018.

\bibitem{Pascanu:2013:DTR:3042817.3043083}
R.~Pascanu, T.~Mikolov, and Y.~Bengio, ``On the difficulty of training
  recurrent neural networks,'' in {\em ICML}, 2013.

\bibitem{6751479}
B.~{Fernando}, A.~{Habrard}, M.~{Sebban}, and T.~{Tuytelaars}, ``Unsupervised
  visual domain adaptation using subspace alignment,'' in {\em 2013 IEEE
  International Conference on Computer Vision}, 2013.

\bibitem{1544935}
J.~{Winn}, A.~{Criminisi}, and T.~{Minka}, ``Object categorization by learned
  universal visual dictionary,'' in {\em Tenth IEEE International Conference on
  Computer Vision (ICCV'05) Volume 1}, 2005.

\bibitem{FridAdar2018SyntheticDA}
M.~Frid-Adar, E.~Klang, M.~Amitai, J.~Goldberger, and H.~Greenspan, ``Synthetic
  data augmentation using gan for improved liver lesion classification,'' in
  {\em ISBI}, 2018.

\bibitem{inproceedings}
M.~Moradi, A.~Madani, A.~Karargyris, and T.~F.~Syeda-Mahmood, ``Chest x-ray
  generation and data augmentation for cardiovascular abnormality
  classification,'' in {\em ISOP}, 2018.

\bibitem{5995347}
A.~{Torralba} and A.~A. {Efros}, ``Unbiased look at dataset bias,'' in {\em
  CVPR 2011}, 2011.

\bibitem{vanDerMaaten2008}
L.~van~der Maaten and G.~Hinton, ``Visualizing data using {t-SNE},'' {\em
  Journal of Machine Learning Research}, 2008.

\bibitem{Redmon2016YouOL}
J.~Redmon, S.~K. Divvala, R.~B. Girshick, and A.~Farhadi, ``You only look once:
  Unified, real-time object detection,'' {\em CVPR}, 2016.

\bibitem{Quionero-Candela:2009:DSM:1462129}
J.~Quionero-Candela, M.~Sugiyama, A.~Schwaighofer, and N.~D. Lawrence, {\em
  Dataset Shift in Machine Learning}.
\newblock 2009.

\bibitem{DBLP:journals/corr/YosinskiCBL14}
J.~Yosinski, J.~Clune, Y.~Bengio, and H.~Lipson, ``How transferable are
  features in deep neural networks?,'' in {\em Advances in Neural Information
  Processing Systems}, 2014.

\bibitem{5995702}
B.~{Kulis}, K.~{Saenko}, and T.~{Darrell}, ``What you saw is not what you get:
  Domain adaptation using asymmetric kernel transforms,'' in {\em CVPR}, 2011.

\bibitem{Saenko:2010:AVC:1888089.1888106}
K.~Saenko, B.~Kulis, M.~Fritz, and T.~Darrell, ``Adapting visual category
  models to new domains,'' in {\em ECCV}, 2010.

\bibitem{6126344}
R.~{Gopalan}, {Ruonan Li}, and R.~{Chellappa}, ``Domain adaptation for object
  recognition: An unsupervised approach,'' in {\em ICCV}, 2011.

\bibitem{6247911}
B.~{Gong}, Y.~{Shi}, F.~{Sha}, and K.~{Grauman}, ``Geodesic flow kernel for
  unsupervised domain adaptation,'' in {\em CVPR}, 2012.

\bibitem{Gong12overcomingdataset}
B.~Gong, F.~Sha, and K.~Grauman, ``Overcoming dataset bias: An unsupervised
  domain adaptation approach,'' in {\em In NIPS Workshop on Large Scale Visual
  Recognition and Retrieval}, 2012.

\bibitem{Ren:2015:FRT:2969239.2969250}
S.~Ren, K.~He, R.~Girshick, and J.~Sun, ``Faster r-cnn: Towards real-time
  object detection with region proposal networks,'' in {\em Proceedings of the
  28th International Conference on Neural Information Processing Systems -
  Volume 1}, 2015.

\bibitem{article}
Y.~Netzer, T.~Wang, A.~Coates, A.~Bissacco, B.~Wu, and A.~Y~Ng, ``Reading
  digits in natural images with unsupervised feature learning,'' 2011.

\bibitem{7797130}
H.~{Zhao}, O.~{Gallo}, I.~{Frosio}, and J.~{Kautz}, ``Loss functions for image
  restoration with neural networks,'' {\em IEEE Transactions on Computational
  Imaging}, 2017.

\bibitem{Saenko}
K.~Saenko, B.~Kulis, M.~Fritz, and T.~Darrell, ``Adapting visual category
  models to new domains,'' in {\em ECCV}, 2010.

\bibitem{Ganin:2016:DTN:2946645.2946704}
Y.~Ganin, E.~Ustinova, H.~Ajakan, P.~Germain, H.~Larochelle, F.~Laviolette,
  M.~Marchand, and V.~Lempitsky, ``Domain-adversarial training of neural
  networks,'' {\em Journal of Machine Learning Research}, 2016.

\bibitem{1467360}
N.~{Dalal} and B.~{Triggs}, ``Histograms of oriented gradients for human
  detection,'' in {\em 2005 IEEE Computer Society Conference on Computer Vision
  and Pattern Recognition (CVPR'05)}, 2005.

\bibitem{Simonyan2013DeepIC}
K.~Simonyan, A.~Vedaldi, and A.~Zisserman, ``Deep inside convolutional
  networks: Visualising image classification models and saliency maps,'' {\em
  CoRR}, vol.~abs/1312.6034, 2013.

\bibitem{Tommasi2015ADL}
T.~Tommasi, N.~Patricia, B.~Caputo, and T.~Tuytelaars, ``A deeper look at
  dataset bias,'' in {\em GCPR}, 2015.

\bibitem{Su2019ActiveAD}
J.-C. Su, Y.-H. Tsai, K.~Sohn, B.~Liu, S.~Maji, and M.~K. Chandraker, ``Active
  adversarial domain adaptation,'' {\em ArXiv}, vol.~abs/1904.07848, 2019.

\bibitem{Kang2018DeepAA}
G.~Kang, L.~Zheng, Y.~Yan, and Y.~Yang, ``Deep adversarial attention alignment
  for unsupervised domain adaptation: The benefit of target expectation
  maximization,'' in {\em ECCV}, 2018.

\bibitem{Long:2015:LTF:3045118.3045130}
M.~Long, Y.~Cao, J.~Wang, and M.~I. Jordan, ``Learning transferable features
  with deep adaptation networks,'' in {\em ICML 37}, 2015.

\bibitem{DBLP:journals/corr/abs-1711-06969}
S.~Sankaranarayanan, Y.~Balaji, A.~Jain, S.~Lim, and R.~Chellappa,
  ``Unsupervised domain adaptation for semantic segmentation with gans,'' {\em
  CVPR}, 2018.

\bibitem{DBLP:journals/corr/LongCWJ17}
M.~Long, Z.~Cao, J.~Wang, and M.~I. Jordan, ``Domain adaptation with randomized
  multilinear adversarial networks,'' {\em Advances in Neural Information
  Processing Systems}, 2018.

\bibitem{726791}
Y.~{Lecun}, L.~{Bottou}, Y.~{Bengio}, and P.~{Haffner}, ``Gradient-based
  learning applied to document recognition,'' {\em Proceedings of the IEEE},
  1998.

\bibitem{1384978}
{Li Fei-Fei}, R.~{Fergus}, and P.~{Perona}, ``Learning generative visual models
  from few training examples: An incremental bayesian approach tested on 101
  object categories,'' in {\em CVPR}, 2004.

\bibitem{DBLP:journals/corr/HeZRS15}
K.~He, X.~Zhang, S.~Ren, and J.~Sun, ``Deep residual learning for image
  recognition,'' 2016.

\bibitem{Kingma2015AdamAM}
D.~P. Kingma and J.~Ba, ``Adam: A method for stochastic optimization,'' 2015.

\bibitem{Ioffe2015BatchNA}
S.~Ioffe and C.~Szegedy, ``Batch normalization: Accelerating deep network
  training by reducing internal covariate shift,'' {\em ICML}, 2015.

\bibitem{Nair:2010:RLU:3104322.3104425}
V.~Nair and G.~E. Hinton, ``Rectified linear units improve restricted boltzmann
  machines,'' in {\em ICML}, 2010.

\bibitem{dumoulin2016guide}
V.~Dumoulin and F.~Visin, ``A guide to convolution arithmetic for deep
  learning,'' {\em ArXiv preprint}, 2018.

\bibitem{imagenet_cvpr09}
J.~Deng, W.~Dong, R.~Socher, L.-J. Li, K.~Li, and L.~Fei-Fei, ``{ImageNet: A
  Large-Scale Hierarchical Image Database},'' in {\em CVPR}, 2009.

\bibitem{5539970}
J.~{Xiao}, J.~{Hays}, K.~A. {Ehinger}, A.~{Oliva}, and A.~{Torralba}, ``Sun
  database: Large-scale scene recognition from abbey to zoo,'' in {\em CVPR},
  2010.

\bibitem{Russell:2008:LDW:1345995.1345999}
B.~C. Russell, A.~Torralba, K.~P. Murphy, and W.~T. Freeman, ``Labelme: A
  database and web-based tool for image annotation,'' in {\em IJCV}, 2008.

\bibitem{Everingham:2010:PVO:1747084.1747104}
M.~Everingham, L.~Gool, C.~K. Williams, J.~Winn, and A.~Zisserman, ``The pascal
  visual object classes (voc) challenge,'' in {\em IJCV}, 2010.

\end{thebibliography}

\end{document}